# Preference Elicitation For General Random Utility Models

**Hossein Azari Soufiani**
SEAS[*], Harvard University
azari@fas.harvard.edu

**David C. Parkes**
SEAS, Harvard University
parkes@eecs.harvard.edu

**Lirong Xia**
SEAS, Harvard University
lxia@seas.harvard.edu

## Abstract

This paper discusses *General Random Utility Models (GRUMs)*. These are a class of parametric models that generate partial ranks over alternatives given attributes of agents and alternatives. We propose two preference elicitation scheme for GRUMs developed from principles in Bayesian experimental design, one for social choice and the other for personalized choice. We couple this with a general Monte-Carlo-Expectation-Maximization (MC-EM) based algorithm for MAP inference under GRUMs. We also prove uni-modality of the likelihood functions for a class of GRUMs. We examine the performance of various criteria by experimental studies, which show that the proposed elicitation scheme increases the precision of estimation.

## 1 Introduction

In many situations, we need to know the preferences of agents over a set of alternatives, in order to make decisions. For example, in recommender systems, we can compute recommendations of new products for a user based on his reported preferences over some products. In social choice, we need to know agent preferences over alternatives, to make a joint decision. Predicting consumer behavior based on reported preferences is an important topic in econometrics [3, 4].

There are two closely related challenges in building a decision support system: preference acquisition and computer-aided decision making [7].

Given preferences, the decision making problem can typically be solved through optimization techniques (e.g., computing the choice that minimizes the maximum *regret*). However, there is often a *preference bottleneck*, where it is too costly or even impossible for users to report full information about their preferences. This happens, for example, in airline recommendation systems, where the number of possible itineraries is large [7]. Another instance is combinatorial voting, where agents vote on multiple related issues [15].

To overcome the preference bottleneck, a well accepted approach is *preference elicitation*. This aims to elicit as little as possible of the agents' preferences, to make a good decision. Previous work focused on achieving one of the following two goals:

1. Social choice. We want to make a joint decision for all agents. Applications include combinatorial auctions [21], voting [11, 17], and crowdsourcing [19].

2. Personalized choice. We want to "learn" an agent's preferences based on a part of her own preferences or preferences of other similar agents. Applications include product configuration [7]. See [5, 13] for recent developments.

In this paper, we focus on elicitation for *ordinal preference*, which means that the agents' preferences are represented by rankings. We assume that preferences are generated by *general random utility models (GRUMs)*. In a GRUM, an agent's preferences are generated as follows: Each alternative is characterized by a *utility distribution*, and the agents rank the alternatives according to the *perceived utilities*, which are generated from the corresponding utility distributions. Parameters for each utility distribution are computed by a combination of attributes of the alternative and attributes of the agent. Parameters of the GRUM model the interrelationship between alternative attributes and agent attributes. See Section 2.1 for more details.

GRUMs are a significant extension of *random utility models (RUMs)* [22], where the effect of attributes of alternatives and agents are not considered. RUMs have been extensively studied and applied in prior work but generally in ways that are specialized to particular parametric forms; e.g., the Bradley-Terry model [8] and the Plackett-Luce model [18, 20].

---
[*]School of Engineering and Applied Sciences.

## 1.1 Contributions

We propose a general adaptive method (Algorithm 1) for preference elicitation within the *Bayesian experimental design* framework (see, [10, e.g.]), guided by maximum expected information gain. In this paper, we focus on a special case, where in each step a targeted agent reports her preferences in full.

We target an agent for elicitation who, based on agent attributes, will provide the greatest expected information gain. In addition to using classical criteria in Bayesian experimental design, we also propose two new criteria that are designed to best improve the quality of the inferred rank preferences, one for predicting social choice, and the other for predicting personalized choice.

Directly computing the optimal agent to target next can be challenging due to the lack of efficient algorithms for MAP inference and lack of efficient computation of observed Fisher information. To overcome this, we extend the MC-EM algorithm and conditions for convergence developed for RUMs by Azari et al. [1] to handle GRUMs. We compute observed Fisher information within the E-step.

We test the prosed methods for MAP/MLE inference and preference elicitation for GRUMs on both synthetic dataset and the Sushi dataset [14].

We compare the performance under the new criteria and performance under the standard criteria from Bayesian experimental design literature. Results show that our elicitation framework can significantly improve the precision of estimation for a moderate number of samples in social choice, relative to random and some of the classical elicitation criteria.

## 1.2 Related Work

GRUMs are a specific case of the generative model studied by Berry, Levinsohn and Pakes (therefore BLP) [3]. The BLP model explicitly considers unobserved attributes of alternatives and agents, whereas GRUMs only consider observed attributes.

However, most work on the BLP model has focused on calculating aggregate properties (for example, the demand curve) when a distribution of the values of unobserved attributes are given. Moreover, the methodologies developed in [3] and subsequent papers only work for the *logit model*. That is: the utility distributions are the standard Gumbel distribution, which is a special case. Even when there are no unobserved variables, BLP was not known to be computationally tractable, beyond the logit case.

An approximate method, that of maximum simulated likelihood has been proposed for GRUMs [23]. We focus on MAP/MLE inference and preference elicitation for GRUMs. We developed an MC-EM algorithm for a large class of GRUMs. To the best of our knowledge, this is the first practical algorithm for MAP/MLE inference for general GRUMs, beyond the logit case. We note that RUMs are a special class of GRUMs. Therefore, the new algorithm naturally extends the algorithm developed by Azari et al. [1] for RUMs. [1]

For social choice, the elicitation scheme designed by Lu and Boutilier [17] aims at computing the outcomes of different commonly studied voting rules. In comparison, the proposed elicitation scheme aims at computing the MAP of GRUMs, which we believe to be different from any commonly studied voting rules.

Compared to the elicitation scheme designed by Pfeiffer et al. [19], which adopted the *Bradley-Terry model*, this paper focuses on GRUMs, which is much more general. Also, as we will see later in the paper in Example 2, the elicitation scheme by Pfeiffer et al. is closely related to a well studied criterion under the Bayesian experimental design framework called *D-optimality*. In contrast, the new elicitation framework allows us to use many other classical criteria in Bayesian experimental design, including D-optimality. Moreover, surprisingly, experimental results on synthetic data show that D-optimality might not be a good choice for social choice for rankings.

The new elicitation framework considers the attributes of agents and alternatives, allowing for more options for elicitation (e.g. we can target an agent with specific attributes). The proposed method is related to the general idea in [13, 9, 6]. However, the proposed method is more general, in the sense that we can handle orders with any length (e.g. Sushi dataset which includes full orders and not only pairwise data). It can also handle any partial order situation due to missing data or design of voting rule (e.g. $k$ first voting or ranks for some missing parties).

## 2 Preliminaries

In this section, we formally define GRUMs and their corresponding MAP mechanism. Further, we recall basic ideas in Bayesian experimental design.

### 2.1 General Random Utility Models

We consider a preference aggregation setting with a set of alternatives $\mathcal{C} = \{c_1, .., c_m\}$, and multiple agents indexed by $i \in \{1, \ldots, n\}$. In GRUMs, for every $j \leq m$, alternative $j$ is characterized by a vector of $L \in \mathbb{M}$ real numbers, denoted by $\vec{z}_j$. And for every $i \leq n$, agent $i$ is characterized by a vector of $K \in \mathbb{N}$ real numbers, denoted by $\vec{x}_i$.[2]

---

[1] Inference and elicitation for GRUMs with unobserved attributes are two interesting directions for future research.

[2] In this paper we focus on the case where all $\vec{x}_i$ and $\vec{z}_j$ are numerical attributes rather than categorical attributes.

Throughout the paper, $j$ denotes an alternative, $i$ denotes an agent, $l$ denotes the attribute of an alternative, and $k$ denotes an agent attribute.

The agents' preferences are generated through the following process.[3] Let $u_{ij}$ be agent $i$'s *perceived utility* for alternative $j$, and let $B$ be a $K \times L$ real matrix that models the linear inter-relation between attributes of alternatives and attributes of agents.

$$u_{ij} = \delta_j + \vec{x}_i B(\vec{z}_j)^T + \epsilon_{ij}, \quad (1)$$
$$u_{ij} \sim \Pr(\cdot | \vec{x}_i, \vec{z}_j, \delta_j, B) \quad (2)$$

In words, agent $i$'s utility for alternative $j$ is composed of the following three parts:

1. $\delta_j$: The *intrinsic utility* of alternative $j$, which is the same across all agents;

2. $\vec{x}_i B(\vec{z}_j)^T$: The *agent-specific utility*, where $B$ is the same across all agents;

3. $\epsilon_{ij}$: The random noise generated independently across agents and alternatives.

Given this, an agent ranks the alternatives according to her perceived utilities for the alternatives in the descending order. That is, for agent $i$, $c_{j_1} \succ_i c_{j_2}$ if and only if $u_{ij_1} > u_{ij_2}$.[4] The parameters for a GRUM are denoted by $\Theta = (\vec{\delta}, B)$. When $K = L = 0$, the GRUM model degenerates to RUM.

**Example 1** Figure 1 illustrates a GRUM for three alternatives (different kinds of sushi) and $n$ agents. Each alternative is characterized by its attributes including heaviness, price, and custom loyalty. Each agent is characterized by attributes including gender and age. Agent attributes have different relationships with alternative attributes. For instance, a person's salary can be related to a preference in regard to the sushi's price rather than heaviness. The outcome of this relationship is a vector of nondeterministic utilities, assigned to the alternatives by each agent.

### 2.2 MAP Inference

Given a GRUM, the preference profile is viewed as *data*, $D = \{\pi^1, \ldots, \pi^n\}$, where each $\pi^i$ is a permutation $(\pi^i(1), \ldots, \pi^i(m))$ of $\{1, \ldots, m\}$ that represents the full ranking $[c_{\pi^i(1)} \succ_i c_{\pi^i(2)} \succ_i \cdots \succ_i c_{\pi^i(m)}]$. We take the standard maximum a posteriori (MAP) approach to estimate the parameters.

Recall that each agent's preferences are generated conditionally independently given the parameters $\Theta$. Therefore, in GRUMs, the probability (likelihood) of the data given

[3]For better presentation, throughout the paper we assume that the preferences are full rankings. The results and algorithms can be extended to the case where the preferences are partial rankings.

[4]For all reasonable GRUMs the situations with tied perceived utilities have zero probability measure.

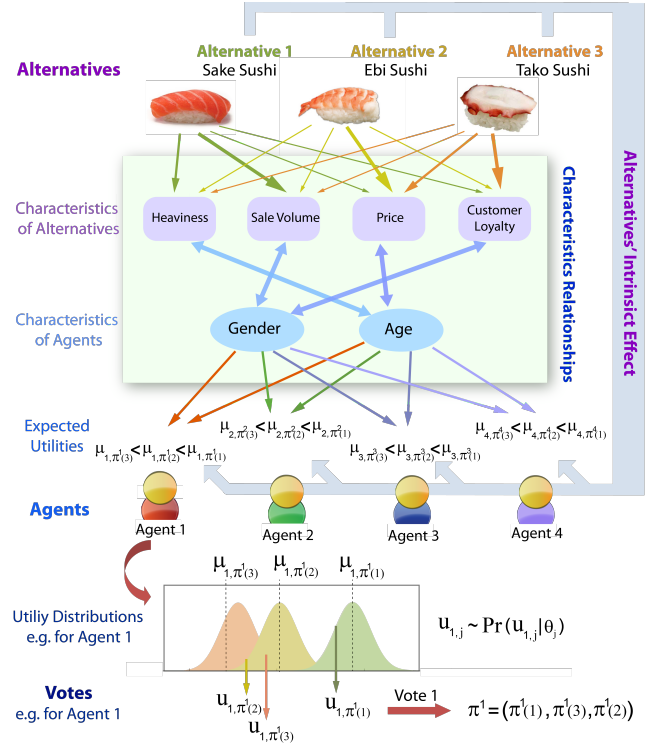

Figure 1: The generative process for GRUMs.

the ground truth $\Theta$ is: $\Pr(D \mid \Theta) = \prod_{i=1}^n \Pr(\pi^i \mid \Theta)$, where:

$$\Pr(\pi^i | \Theta) = \int_{u_{i\pi^i(1)} > \cdots > u_{i\pi^i(n)}} \prod_j \Pr(u_{i\pi^i(j)} | \vec{x}_i, \vec{z}_j, \Theta)\, du_{i\pi^i(j)}$$

Suppose we have a prior over the parameters, for MAP inference we aim at computing $\Theta$ to maximize the posterior function:

$$\Pr(\Theta | D) = \prod_{i=1}^n \Pr(\pi^i \mid \Theta) \Pr(\Theta)$$

After computing $\Theta^*$ that maximizes posterior, we can make joint decisions for the agents based on $\Theta^*$.[5] For example, we can choose the winner to be the alternative whose utility distribution has the highest mean, or choose a winning ranking over alternatives by ranking the means of the utility distributions.

### 2.3 One-Step Bayesian Experimental Design

Suppose we have a parametric probabilistic model. Let $\Pr(\Theta^*)$ denote the prior distribution over the parameters. A one-step Bayesian experimental design problem is composed of two parts: a set of *designs* $\mathcal{H}$ and a *quality func-*

[5]In the context of social choice, the prior is often uniform, and MAP becomes MLE.

tion $G(\cdot)$ defined on any *distribution* over the parametric space.

A design $h \in \mathcal{H}$ is mathematically characterized by $\Pr(\cdot|\Theta^*, h)$ that controls the way the data $D$ are generated for any ground truth parameter vector $\Theta^*$. Therefore, for any given design $h$, we can compute the probability for data $D$ as $\Pr(D|h)$. Given any data $D$ and design $h$, we can compute the posterior distribution of parameters $\Pr(\cdot|D, h)$. The objective of Bayesian experimental design is to choose the design $h$ that maximizes the expected quality of the posterior of MAP parameters, where the randomness comes from the data that are generated given $h$. Formally, we aim at computing $h^*$ as follows.

$$h^* = \arg\max_h \int G(\Pr(\cdot|D, h)) \times \Pr(D|h) \, dD \quad (3)$$

Often, directly computing (3) is hard. Even $G(\Pr(\cdot|D, h))$ is difficult to compute given $D$ and $h$. Researchers have taken various approximations to (3). A common approach is to approximate $\Pr(\cdot|D, h)$ by a normal distribution $\mathcal{N}(\hat{\Theta}, [R(\hat{\Theta}) + I_h(\hat{\Theta})]^{-1})$, where:

- $\hat{\Theta}$ is the MAP of $D$,

- $R(\Theta)$ is the *precision matrix* of the prior over $\Theta$, that is, $R = \nabla^2_\Theta \log \Pr(\Theta)$, and

- $I_h(\hat{\Theta})$ is the *Fisher information* matrix defined as follows. Let $X_\pi = \nabla_\Theta \log \Pr(\pi|\vec{\Theta}, h)$, we have

$$I_h(\hat{\Theta}) = E_\pi(X_\pi(X_\pi)^T|_{\Theta=\hat{\Theta}}).$$

Equivalently, if $\log \Pr(\pi|\Theta, h)$ is twice differentiable w.r.t. $\Theta$ for each ranking $\pi$, then

$$I_h(\hat{\Theta}) = -E_\pi(\nabla^2_\Theta \log \Pr(\pi|\Theta, h)|_{\Theta=\hat{\Theta}}).$$

If we approximate $\Pr(\cdot|D, h)$ by $\mathcal{N}(\hat{\Theta}, [R(\hat{\Theta}) + I_h(\hat{\Theta})]^{-1})$, then the most commonly studied quality functions are functions of $\hat{\Theta}$ and $h$. More precisely, they are functions of $\hat{\Theta}$ and $R(\hat{\Theta}) + I_h(\hat{\Theta})$. In such cases, we can rewrite $G(\mathcal{N}(\hat{\Theta}, I_h(\hat{\Theta}))) = G^*_R(\hat{\Theta}, h)$. Then, (3) becomes:

$$h^* = \arg\max_h \int G^*_R(\hat{\Theta}, h) \cdot \Pr(\hat{\Theta}|h) d\hat{\Theta} \quad (4)$$

Still the integration in (4) is often hard to compute, and is approximated by $G^*_R(\Theta^*, h)$, where $\Theta^*$ is the mode of $\Pr(\Theta)$. Some popular quality functions and corresponding approximations are summarized in Table 1.

**Example 2** *The adaptive elicitation approach by Pfeiffer et al. [19] is a special case of Bayesian D-optimality design, where $\mathcal{H}$ is the set of all pairwise questions between alternatives. Pfeiffer et al. derived formulas for $\Pr(\cdot|\Theta^*, h)$ for each $h \in \mathcal{H}$, and chose $h^*$ according to (3). The quality function they use is the negative Shannon entropy, which is exactly D-optimality as shown in Table 1.*

## 3 Our Preference Elicitation Scheme

In the new elicitation framework, we adapt the one-step Bayesian experimental design to multiple iterations. For any iteration $t$, let $D^t$ denote the preferences elicited in all previous iterations. The prior distribution $\Pr^t$ over parameters is the posterior of observing $D^t$, that is: for any $\Theta$, $\Pr^t(\Theta) = \Pr(\Theta|D^t)$. Then we solve a standard one-step Bayesian experimental design problem w.r.t. the prior $\Pr^t$ to elicit a new agents' preferences, and then form $D^{t+1}$ for the next iteration.

Our general elicitation framework for GRUMs is presented as Algorithm 1. To allow flexibility of using various criteria of Bayesian experimental design, we let the input consist of the heuristic $G^*_R(\hat{\Theta}, h)$, which is usually a function of $\hat{\Theta}$ and $R(\hat{\Theta}) + I_h(\hat{\Theta})$. To present the main idea, in this paper the set of designs $\mathcal{H}$ is the multi-set of all agents attributes. That is, in each iteration (Steps 1~3) we will compute an $h \in \mathcal{H}$ and query the preferences of a random agent whose attributes are $h$.[6] Steps 1~3 are hard to

---
**Algorithm 1** Preference Elicitation for GRUMs

**Heuristic:** $G^*_R(\hat{\Theta}, h)$.
Randomly choose an initial set of data $D^1$.
**for** $t = 1$ to $T$ **do**
   **1:** Compute $\Theta^t = \text{MAP}(D^t)$.
   **2:** Compute the precision matrix $R^t$ of $\Pr(\Theta|D^t)$ at $\Theta^t$.
   **3:** Compute $h^t \in \mathcal{H}$ that maximizes $G^*_{R^t}(\Theta^t, h^t)$.
   **4:** Query an agent whose attributes are $h^t$. Let $\pi^t$ denote her preferences. $D^{t+1} \leftarrow D^t \cup \{\pi^t\}$, $\mathcal{H} \leftarrow \mathcal{H} \setminus \{h^t\}$.
**end for**

---

compute. In this paper, we will use a multivariate normal distribution $\mathcal{N}(\hat{\Theta}, J_{D^t}(\hat{\Theta})^{-1})$ to approximate $\Pr(\Theta|D^t)$ in Step 2, where $J_{D^t}(\hat{\Theta})$ is the *observed Fisher information* matrix, and we immediately have $R^t = J_{D^t}(\hat{\Theta})$.[7] Given any data $D$, $J_D(\hat{\Theta}, h)$ is defined as follows. Again, let $\hat{\Theta} = \text{MAP}(D)$.

$$J_{D,h}(\hat{\Theta}) = \sum_{\pi \in D} (X_\pi \times (X_\pi)^T|_{\Theta=\hat{\Theta}}).$$

Equivalently, if $\log \Pr(\pi|\Theta, h)$ is twice differentiable w.r.t. $\Theta$ for each ranking $\pi$, then we have:

$$J_{D,h}(\hat{\Theta}) = -\sum_{\pi \in D} (\nabla^2_\Theta \log \Pr(\pi|\Theta, h)|_{\Theta=\hat{\Theta}}).$$

In Section 4 we propose an MC-EM algorithm to compute $\text{MAP}(D^t)$ in Step 1. In Section 4.3 we study how to compute the observed Fisher information matrix $R^t = J_{D^t}(\Theta^t)$, and use it for elicitation as well as accelerating MC-EC algorithm. Computation of the Fisher information matrix $I_h(\hat{\Theta})$ used in Step 3 will also be discussed in Section 4.3.

---
[6]The elicitation scheme can be extended to other types of elicitation questions, for instance, pairwise comparisons and "top-$k$".
[7]See e.g. page 224 [2] for justification of this approximation.

| Name | Quality function | Heuristics $G_R^*(\hat{\Theta}, h)$ |
|---|---|---|
| D-optimality | Gain in Shannon information | $det(R + I_h(\hat{\Theta}))$ |
| E-optimality | Minimum eigenvalue of the information matrix | $\lambda_{min}\{R + I_h(\hat{\Theta})\}$ |
| Proposed criterion for social choice | Minimum inverse of pairwise coefficient of variation | Equation (5) |
| Proposed criterion for personalized choice | Minimum inverse of pairwise coefficient of variation | Equation (6) |

Table 1: Different criteria for experimental design.

The choice of $G_R^*$ is crucial for the performance of the elicitation algorithm. The two first criteria summarized in Table 1 are generic criteria for making the posterior as certain as possible, which may not work well for eliciting the aggregated ranking or individual rankings. In Section 6 we report experimental results comparing performance of different $G_R^*$ in Table 1 and the new criteria we propose for both social choice and individual ranking.

### 3.1 A New Elicitation Criterion for Social Choice

The social choice ranking is the ranking over the components of $\vec{\delta}$. Therefore, if the objective is to elicit preferences for the aggregated ranking, it makes sense to make each pairwise comparison as certain as possible. Following the idea in t-test, we propose to use $\frac{|\text{mean}(\delta_{j_1} - \delta_{j_2})|}{\text{std}(\delta_{j_1} - \delta_{j_2})}$ (which is the inverse of *coefficient of variation*) to evaluate the certainty in pairwise comparison between $c_{j_1}$ and $c_{j_2}$. The larger the value is, the more certain we are about the comparison between $c_{j_1}$ and $c_{j_2}$. Therefore, we propose to use the following quality function $G$ distributions over $\Theta$. We recall that $\Theta = (\vec{\delta}, B)$.

$$G(\text{Pr}) = \min_{j_1 \neq j_2} \frac{|\text{mean}(\delta_{j_1} - \delta_{j_2})|}{\text{std}(\delta_{j_1} - \delta_{j_2})}.$$

In words, $G$ is the minimum inverse of the coefficient of variation across all pairwise comparisons. The corresponding $G_R^*$ is thus the following.

$$G_R^*(\Theta, h) = \min_{j_1 \neq j_2} \frac{|\text{mean}(\delta_{j_1} - \delta_{j_2})|}{\sqrt{\text{Var}(\delta_{j_1}) + \text{Var}(\delta_{j_2}) + 2\text{cov}(\delta_{j_1}, \delta_{j_2})}}, \quad (5)$$

Where $|\text{mean}(\delta_{j_1} - \delta_{j_2})|$ can be computed from $\Theta$ and $\sqrt{\text{Var}(\delta_{j_1}) + \text{Var}(\delta_{j_2}) + 2\text{cov}(\delta_{j_1}, \delta_{j_2})}$ can be computed from $R + I_h(\Theta)$.

### 3.2 Generalization to Personalized Choice

Following the idea in the new criterion proposed in the last subsection for social choice, for any agent with attributes $\vec{x}$, we can define a similar quality function $G_{\vec{x}}(\text{Pr})$. This makes the ranking of the alternatives w.r.t. the deterministic parts of the perceived utilities[8] as certain as possible, as follows. For any $j \leq m$, let $\mu_j = \delta_j + \vec{x}B(\vec{z}_j)^T$. We note that $\mu_j$ is a linear combination of the parameters in $\Theta$.

$$G_{\vec{x}}(\text{Pr}) = \min_{j_1 \neq j_2} \frac{|\text{mean}(\mu_{j_1} - \mu_{j_2})|}{\text{std}(\mu_{j_1} - \mu_{j_2})} \quad (6)$$

$G_{\vec{x}}^*(\Theta, h)$ can be defined in a similar way. However, usually we want to predict the rankings for a population of agents, for which only a distribution over agent attributes is known. Mathematically, let $\Delta$ denote a probability distribution over $\mathbb{R}^L$. We can extend the criterion for personalized choice w.r.t. $\Delta$ as follows.

$$G_\Delta(\text{Pr}) = \int_{\vec{x} \in \mathbb{R}^T} G_{\vec{x}}(\text{Pr}) \cdot \Delta(\vec{x}) \, d\vec{x}.$$

$G_\Delta$ is usually hard to compute since it involves integrating $G_{\vec{x}}$ over all $\vec{x}$ in support of $\Delta$, which is often not analytically or computationally tractable. In the experiments, we will use the criterion defined in (5) for personalized ranking and surprisingly it works well.

## 4 An MC-EM Inference Algorithm

In this section, we extend MC-EM algorithm for RUMs proposed by Azari et al. [1] to GRUMs. We focus on GRUMs where the conditional probability $\text{Pr}(\cdot|\vec{x}_i, \vec{z}_j, \delta_j, B)$ belongs to the *exponential family*, which takes the following form: $\text{Pr}(U = u|\vec{x}_i, \vec{z}_j, \delta_j, B) = e^{\eta_{ij} \cdot T(u) - A(\eta_{ij}) + H(u)}$, where $\eta_{ij}$ is the vector of *natural parameters*, which is a function of $\vec{x}_i, \vec{z}_j, \Theta$. $A$ is a function of $\eta_{ij}$ and $T$ and $H$ are functions of $u$.

Let $U = (\vec{u_1}, \ldots, \vec{u_n})$ denote the latent space, where $\vec{u_i} = (u_{i1}, \ldots, u_{im})$ represent agent $i$'s perceived utilities for the alternatives. The general framework of the proposed EM algorithm is illustrated in Algorithm 2. The algorithm has multiple iterations, and in each iteration there is an E-step and a general M-step. Therefore, the algorithm is a *general EM (GEM)* algorithm. We recall that $\Theta = (\vec{\delta}, B)$ represents the parameters.

The algorithm is performed for a fixed number of iterations or until no $\Theta^{t+1}$ in the M-step can be found. However, the

---
[8]That is, the intrinsic utility plus personalized utility.

**Algorithm 2** Framework of the EM algorithm

In each iteration.

**E-Step :** $Q(\Theta, \Theta^t)$
$$= E_{\vec{U}} \left\{ \log \prod_{i=1}^{n} \Pr(\vec{u}_i, \pi^i | \Theta) + \log(\Pr(\Theta)) | D, \Theta^t \right\} \quad (7)$$

**M-step :** compute $\Theta^{t+1}$ s.t. $Q(\Theta^{t+1}, \Theta^t) > Q(\Theta^t, \Theta^t)$

---

E-step cannot be done analytically in general, and we will use a Monte Carlo approximation for the E-step.

### 4.1 Monte Carlo E-Step: Gibbs Sampling

Our E-step is similar to the E-step in [1] with a modification that considers the prior. We recall that $\Pr(\cdot | \vec{x}_i, \vec{z}_j, \delta_j, B)$ belongs to the exponential family. We have the following calculation for iteration $t$, where $\mu_{ij} = \delta_{ij} + \vec{x}_i B(\vec{z}_j)^T$ for any given $\Theta = (\vec{\delta}, B)$, and $\mu_{ij}^t = \delta_{ij}^t + \vec{x}_j B^t(\vec{z}_i)^T$.

$$Q(\Theta, \Theta^t) = E_{\vec{U}} \{ \log \prod_{i=1}^{n} \Pr(\vec{U}_i, \pi^i | \Theta) + \log \Pr(\Theta) | D, \Theta^t \}$$

$$= \sum_{i=1}^{n} \sum_{j=1}^{m} E_{u_{ij}} \{ \log \Pr(u_{ij} | \Theta) | \pi^i, \Theta^t \}$$

$$= \sum_{i=1}^{n} \sum_{j=1}^{m} \eta_{ij} S_{ij}^t - A(\eta_{ij}) + W,$$

$$\text{where} \quad S_{ij}^t = E_{u_{ij} \sim \Pr(u_{ij} | \eta_{ij}^t)} \{ u_{ij} | \pi^i \}. \quad (8)$$

We use a Monte Carlo approximation similar to that used in [1], which involves sampling $U$ from the distribution $\Pr(U | D, \Theta^t)$ using a Gibbs sampler, and then approximate $S_{ij}^{t+1}$ by $\frac{1}{N} \sum_{k=1}^{N} u_{ij}^k$. Each step of the Gibbs sampler is sampling from a truncated exponential distribution, illustrated in Figure 2 in [1].

### 4.2 General M-Step

After we compute $S_{ij}^{t+1}$'s, the M-step aims at improving $Q(\Theta, \Theta^t)$:

$$Q(\Theta, \Theta^t) = \sum_{j=1}^{m} \sum_{i=1}^{n} \log \Pr_j(u_{ij} = S_{ij}^{t+1} | \Theta) + \log(\Pr(\Theta))$$

We use steps of Newton's method to improve $Q(\Theta, \Theta^t)$ in the M-step (we can use as many steps at each iteration to ensure the convergence for each M-step).

$$\Theta^{t+1} = \Theta^t - (\nabla_\Theta^2 Q(\Theta, \Theta^t)|_{\Theta^t})^{-1} \nabla_\Theta Q(\Theta, \Theta^t)|_{\Theta^t} \quad (9)$$

$\nabla_\Theta^2 Q(\Theta, \Theta^t)$ and $\nabla_\Theta Q(\Theta, \Theta^t)$ can be computed immediately from $S_{ij}^t$ as follows.

$$\nabla_\Theta^2 Q(\Theta, \Theta^t) = \sum_{i=1}^{n} \sum_{j=1}^{m} \nabla_\Theta^2 \eta_{ij} S_{ij}^t - \nabla_\Theta^2 A(\eta_{ij})$$

$$\nabla_\Theta Q(\Theta, \Theta^t) = \sum_{i=1}^{n} \sum_{j=1}^{m} \nabla_\Theta \eta_{ij} S_{ij}^t - \nabla_\Theta A(\eta_{ij})$$

### 4.3 Computing Observed Fisher information

Computation of the observed Fisher information will not only be used in Step 2 of the new elicitation scheme Algorithm 1, but also will accelerate the GEM algorithm [16]. Fisher information can be computed by the following method proposed by Louis [16]. From the independence of agents we have: $J_D(\hat{\Theta}) = \sum_i J_{\pi^i}(\hat{\Theta})$, where,

$$J_{\pi^i}(\Theta) = E_{U_i} \{ -\nabla_\Theta^2 \log P(\pi^i, U_i | \Theta) | \Theta, \pi^i \}$$
$$- E_{U_i} \{ \nabla_\Theta \log P(\pi^i, U_i | \Theta) \nabla_\Theta \log P(\pi^i, U_i | \Theta)^T | \Theta, \pi^i \}$$

$J_{\pi^i}(\hat{\Theta})$ is computed using the samples ($u_{ij}$'s) generated in MC step in every iteration of EM algorithm as follows.

$$\nabla_\Theta^2 \log P(\pi^i, U_i | \Theta) = \sum_{i=1}^{n} \sum_{j=1}^{m} \nabla_\Theta^2 \eta_{ij} U_{ij} - \nabla_\Theta^2 A(\eta_{ij})$$

$$\nabla_\Theta \log P(\pi^i, U_i | \Theta) = \sum_{i=1}^{n} \sum_{j=1}^{m} \nabla_\Theta \eta_{ij} U_{ij} - \nabla_\Theta A(\eta_{ij})$$

The Fisher information matrix $I_h(\hat{\Theta})$ used in Step 3 of Algorithm 1 can be approximated by $\lim_{n \to \infty} \frac{J_{D_n}(\hat{\Theta})}{n}$, where $D_n$ is the dataset of $n$ rankings randomly generated according to $\Pr(\pi | \hat{\Theta})$. Therefore, we can use the techniques developed in this subsection to approximately compute $I_h(\hat{\Theta})$.

### 4.4 MC-EM Algorithm in Detail

The details of the proposed EM algorithm (with fixed number of iterations) are illustrated in Algorithm 3.

---

**Algorithm 3** MAP for GRUM

**Input:** $D = (\pi^1, \ldots, \pi^n), \Theta^{start}, T \in \mathbb{N}$
Let $\Theta^0 = \Theta^{start}$
**for** $t = 1$ to $T$ **do**
    **for** every $\pi^i \in D$ **do**
        Compute $S_{ij}^{t+1}$ and $J(\Theta^{t+1})$ according to (8) for all $j \leq m$.
    **end for**
    Compute $\Theta^{t+1}$ according to (9).
**end for**

# 5 Global Optimality for Posterior Distribution

In this section, we generalize theorems on global optimality of likelihood for RUMs proved in [1] to GRUMs. All proofs are omitted due to the space constraint. The EM algorithm tends to find local optimal of the posterior distribution, hence, proving global optimality of MAP helps to avoid issues due to EM. First, we present concavity of the posterior distribution in GRUMs.

**Theorem 1** *For the location family, if for every $j \leq m$ the joint probability density function for $\vec{\epsilon_i}$ and the prior $\Pr(\Theta)$ are log-concave, then $\Pr(\Theta|D)$ is concave up to a known transformation.*

For P-L, Ford, Jr. [12] proposed the following necessary and sufficient condition for the set of global maxima solutions to be bounded (more precisely, unique) when $\sum_{j=1}^{m} e^{\Theta_j} = 1$. The conditions are generalized to the case of RUMs in [1]. We prove that this condition is also necessary and sufficient for global maxima solutions of the likelihood function of GRUMS to be bounded.

**Condition 1** *Given the data $D$, in every partition of the alternatives $\mathcal{C}$ into two nonempty subsets $\mathcal{C}_1 \cup \mathcal{C}_2$, there exists $c_1 \in \mathcal{C}_1$ and $c_2 \in \mathcal{C}_2$ such that there is at least one ranking in $D$ where $c_1 \succ c_2$.*

**Theorem 2** *Suppose we fix $\mu_{11} = 0$. Then, the set $S_D$ of global maxima solutions to $\Pr(\Theta|D)$ is bounded in $\Theta$ if and only if the data $D$ satisfies Condition 1 and the linear model describing $\mu$ in terms of $\Theta$ is identifiable.*

# 6 Experimental Results

In this section, we report experimental results on synthetic data and a Sushi dataset from Kamishima [14] for three types of tests described below.

## 6.1 Social Choice and Synthetic Data

We first show the consistency of the model for social choice. We generate random data sets with $\delta_j \sim \text{Normal}(1,1), B_{ij} \sim \text{Normal}(0,1), X_i \sim \text{Normal}(0,1), Z_i \sim \text{Normal}(0,1)$, and then generate random utilities with the random noise $\epsilon_{ij}$ generated with mean zero and variance of 1. The results in Figure 2 are generated by varying the number of agents for which we have preference information. For each number of agents, we estimate the parameter set $\Theta$, and evaluate the Kendall correlation between estimated and true ranks with respect to $\delta_j$'s. These results illustrate the improvement in estimated social choice order as the number of agents in the population increases.

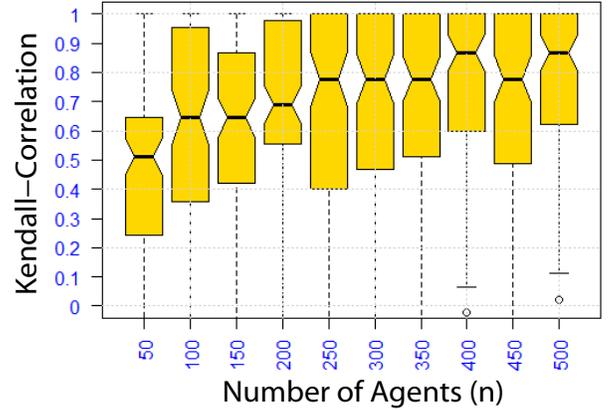

Figure 2: Asymptotic behavior for synthetic data and social choice. The $y$-axis is the average Kendal correlation between the estimated social choice and the ground truth order.

In studying elicitation for social choice, we test the performance of the elicitation schemes shown in Table 1, i.e. D-optimality, E-optimality, and the proposed criterion in (5), and compare the results to random elicitation. We adopt the following two synthetic datasets:

**Dataset 1:** $(B_{ij} \sim N(0,1), X_i \sim N(0,1), Z_i \sim N(0,1)), \delta_j \sim 0.1 * N(1,1)$ and the error term $\epsilon_{ij} \sim N(0,1)$.

**Dataset 2:** The same as Dataset 1, except that the $\delta_j \sim N(1,1)$ and the error term $\epsilon_{ij} \sim N(0, 1/4)$.

Compared to the GRUM in Dataset 1, the model adopted in Dataset 2 has a heavier social component and less noise. For each dataset we generate 100 agents' preferences, and use the three criteria shown in Table 1 to elicit $n \in [1, 100]$ rankings. For each $n$, we apply Algorithm 3 and compare the ranking over the learned $\delta_j$'s with the ground truth social choice ranking.

The results are shown in Figure 3 (graphs are smoothed with a moving window with length 25), where the $x$-axis is the number of agents whose preferences are elicited, and the $y$-axis is the Kendall correlation between the learned ranking and the ground truth ranking. We make the following observations.

• In Dataset 1 where the social component is small, it is not clear which criteria is better, as shown in Figure 3(a), and there are no statistically significant results.

• In Dataset 2 where the social component is large, E-optimality generally works better than the proposed method, while both work better than random, which works surprisingly better than D-optimality, as shown in Figure 3(b). However, only a few of these observations are statistically significant with 90% confidence, for example, considering the interval of $[34, 44]$ agents, E-optimality and the proposed method outperforms Random but the comparison between the other methods is not significant at 90%.

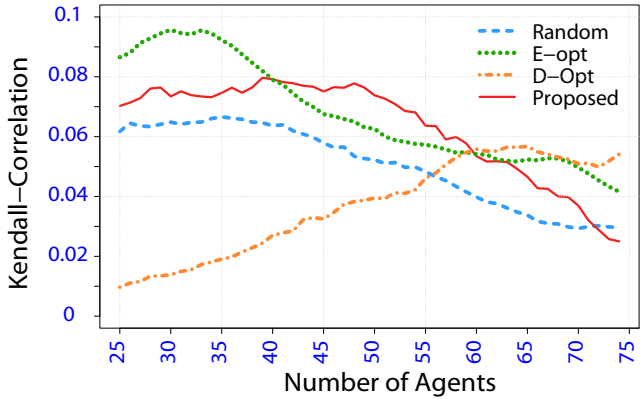

(a) Social choice: Dataset 1.

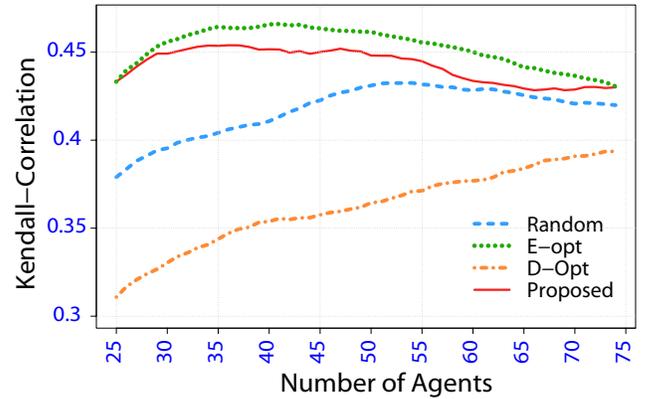

(b) Social choice: Dataset 2.

Figure 3: Comparison of elicitation criteria described in Table 1 for synthetic data and social choice.

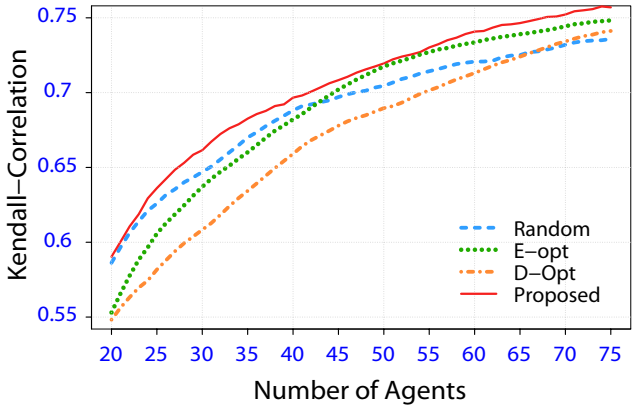

(a) Personalized choice: Dataset 1.

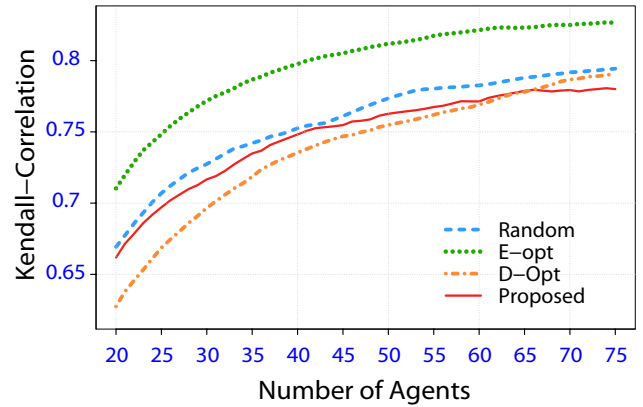

(b) Personalized choice: Dataset 2.

Figure 4: Comparison of elicitation criteria described in Table 1 for synthetic data for personalized choice.

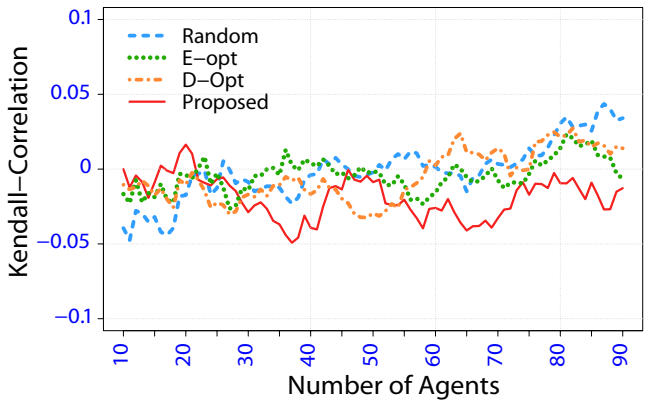

(a) Social choice: Sushi dataset.

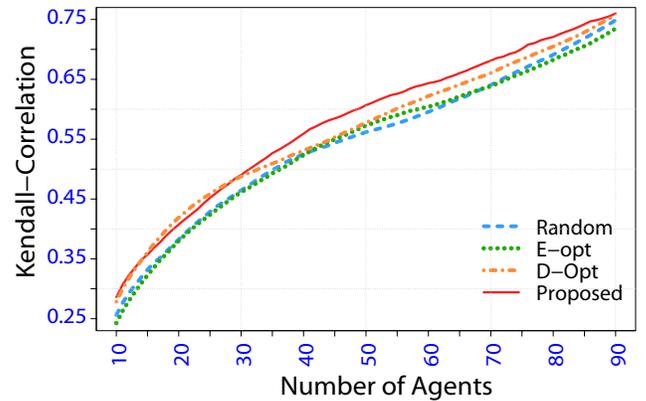

(b) Personalized choice: Sushi dataset.

Figure 5: Comparison of elicitation criteria described in Table 1 for the Sushi dataset [14].

### 6.2 Personalized Choice and Synthetic Data

For personalized choice we first show the consistency results in Figure 6, where the bottom box-plot shows the Kendall correlation between noisy data (i.e., an individual agent's random utility and thus preference order) and the true preference order for each agent, and the top box-plot shows Kendall correlation between estimated agent preference orders and true preference orders, as obtained through the model.

Turning to preference elicitation, we compare the schemes in Table 1 with the random method for the same two datasets as were adopted for social choice. The results are shown in Figure 4(graphs are smoothed with a moving window with length 20). For each group of 100 agents, and for any $n \in [1, 100]$ and each elicitation scheme, we compute the MAP of $\Theta$, and use it to compute the Kendall correlation between the true preferences and the predicted preference for all 100 agents in this group. We make the following observations:

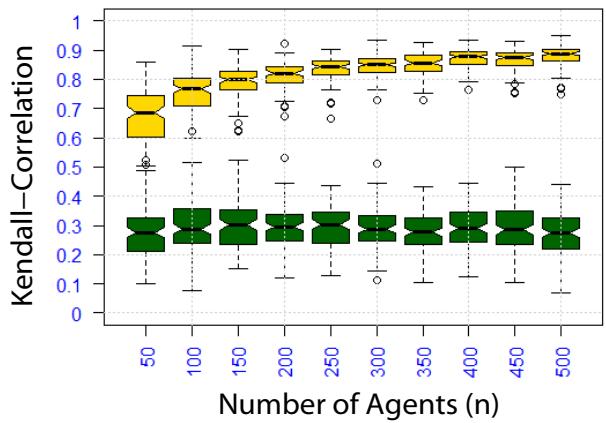

Figure 6: Asymptotic behavior for synthetic data and personalized choice. The $y$-axis is the average Kendall correlation between an estimated preference order and ground truth preference order for an agent. The top box-plot shows the result of inference, the bottom box-plot the correlation from raw data.

• In Dataset 1, where the social component is small, when the number of agents used in elicitation is not too large ($<$ 50), the proposed method works better than E-optimality, which is itself comparable to random. Both methods are better than D-optimality. See Figure 4(a). Some of these observations are statistically significant, for example, when $n = [24, 25]$, E-optimality works better than D-optimality with 90% significance, E-optimality works better than random with 75% significance, the proposed method works better than E-optimality with 75% significance, and the proposed method works better than D-optimality with 75% significance.

• In Dataset 2, where the social component is large, E-optimality generally works better than the proposed method, both work better than random, and random is more effective than D-optimality, as shown in Figure 4(b). However, only a few of these observations are statistically significant with 90% confidence interval, for example E-optimality outperforms D-optimality when the number of agents is in the interval $[29, 42]$.

### 6.3 Sushi Data

In synthetic experiments, we have access to the ground truth. However, in the real world data (Sushi data) there are no data available as ground truth. In this experiment, we estimated parameters $\Theta$ using preferences from 1000 agents, randomly chosen from the 5000 agents in the dataset. And adopt those parameters as the ground truth for the experimental study. The categorical features are discarded from the data set.[9]

The results are shown in Figure 5 (graphs are smoothed with a moving window with length $10$), where (a) shows comparisons for social choice (where we rank $\delta$'s), and (b) shows comparisons for personalized choice. We make the following observations:

• For social choice (a), none of the criteria work well (and note that the Kendall correlations are low for all criteria). We feel that this is reasonable since preferences over sushi is likely high personalized with a small social component to preferences.

• For personalized choice (b), we observe that the proposed method is generally the most effective, while the performance of E-optimality and D-optimality is very close to random. None of these results are statistically significant with $90\%$ confidence.

## 7 Conclusion and Future Work

We have proposed a method for preference elicitation based on ordinal rank data, adopting the framework of Bayesian experimental design. This includes two new criteria for social and personalized case. The proposed criterion for social choice can significantly improve the precision of estimation, relative to random and some of the classical elicitation criteria. This work can also be seen as preference elicitation for learning to rank. In the future, we can adopt the methodology in other preference elicitation applications; for example recommendation systems, product prediction and so forth.

## Acknowledgments

This work is supported in part by NSF Grant No. CCF-0915016. Lirong Xia is supported by NSF under Grant #1136996 to the Computing Research Association for the CIFellows Project. We thank Simon Lunagomez for providing valuable comments on the Bayesian experimental design aspect of this work. We also thank Edoardo Airoldi, Craig Boutilier, Jonathan Huang, Tyler Lu, Nicolaus Tideman, Paolo Viappiani, and anonymous UAI-13 reviewers, for helpful comments and suggestions, or help on the datasets.

---

[9]We focus on non categorical features in this work. The method can be extended to categorical features.

# References

[1] Hossein Azari Soufiani, David C. Parkes, and Lirong Xia. Random utility theory for social choice. In *Proceedings of the Annual Conference on Neural Information Processing Systems (NIPS)*, pages 126–134, Lake Tahoe, NV, USA, 2012.

[2] James O. Berger. *Statistical Decision Theory and Bayesian Analysis*. James O. Berger, 2nd edition, 1985.

[3] Steven Berry, James Levinsohn, and Ariel Pakes. Automobile prices in market equilibrium. *Econometrica*, 63(4):841–890, 1995.

[4] Steven Berry, James Levinsohn, and Ariel Pakes. Differentiated products demand systems from a combination of micro and macro data: The new car market. *Journal of Political Economy*, 112(1):68–105, 2004.

[5] Edwin Bonilla, Shengbo Guo, and Scott Sanner. Gaussian process preference elicitation. In *Advances in Neural Information Processing Systems 23*, pages 262–270. 2010.

[6] Craig Boutilier. On the foundations of expected expected utility. In *Proceedings of the Eighteenth International Joint Conference on Artificial Intelligence (IJCAI)*, pages 285–290, Acapulco, Mexico, 2003.

[7] Craig Boutilier. Computational Decision Support: Regret-based Models for Optimization and Preference Elicitation. In P. H. Crowley and T. R. Zentall, editors, *Comparative Decision Making: Analysis and Support Across Disciplines and Applications*. Oxford University Press, 2013.

[8] Ralph Allan Bradley and Milton E. Terry. Rank analysis of incomplete block designs: I. The method of paired comparisons. *Biometrika*, 39(3/4):324–345, 1952.

[9] Urszula Chajewska, Daphne Koller, and Ron Parr. Making rational decisions using adaptive utility elicitation. In *Proceedings of the National Conference on Artificial Intelligence (AAAI)*, pages 363–369, Austin, TX, USA, 2000.

[10] Kathryn Chaloner and Isabella Verdinelli. Bayesian Experimental Design: A Review. *Statistical Science*, 10(3):273—304, 1995.

[11] Vincent Conitzer and Tuomas Sandholm. Vote elicitation: Complexity and strategy-proofness. In *Proceedings of the National Conference on Artificial Intelligence (AAAI)*, pages 392–397, Edmonton, AB, Canada, 2002.

[12] Lester R. Ford, Jr. Solution of a ranking problem from binary comparisons. *The American Mathematical Monthly*, 64(8):28–33, 1957.

[13] Neil Houlsby, Jose Miguel Hernandez-Lobato, Ferenc Huszar, and Zoubin Ghahramani. Collaborative gaussian processes for preference learning. In *Proceedings of the Annual Conference on Neural Information Processing Systems (NIPS)*, pages 2105–2113. Lake Tahoe, NV, USA, 2012.

[14] Toshihiro Kamishima. Nantonac collaborative filtering: Recommendation based on order responses. In *Proceedings of the Ninth International Conference on Knowledge Discovery and Data Mining (KDD)*, pages 583–588, Washington, DC, USA, 2003.

[15] Jérôme Lang and Lirong Xia. Sequential composition of voting rules in multi-issue domains. *Mathematical Social Sciences*, 57(3):304–324, 2009.

[16] Thomas A. Louis. Finding the observed information matrix when using the EM algorithm. *Journal of the Royal Statistical Society Series B (Statistical Methodology)*, 44:226–233, 1982.

[17] Tyler Lu and Craig Boutilier. Robust approximation and incremental elicitation in voting protocols. In *Proceedings of the Twenty-Second International Joint Conference on Artificial Intelligence (IJCAI)*, pages 287–293, Barcelona, Catalonia, Spain, 2011.

[18] Robert Duncan Luce. *Individual Choice Behavior: A Theoretical Analysis*. Wiley, 1959.

[19] Thomas Pfeiffer, Xi Alice Gao, Andrew Mao, Yiling Chen, and David G. Rand. Adaptive Polling and Information Aggregation. In *Proceedings of the National Conference on Artificial Intelligence (AAAI)*, pages 122–128, Toronto, Canada, 2012.

[20] Robin L. Plackett. The analysis of permutations. *Journal of the Royal Statistical Society. Series C (Applied Statistics)*, 24(2):193–202, 1975.

[21] Tuomas Sandholm and Craig Boutilier. Preference elicitation in combinatorial auctions. In Peter Cramton, Yoav Shoham, and Richard Steinberg, editors, *Combinatorial Auctions*, chapter 10, pages 233–263. MIT Press, 2006.

[22] Louis Leon Thurstone. A law of comparative judgement. *Psychological Review*, 34(4):273–286, 1927.

[23] Joan Walker and Moshe Ben-Akiva. Generalized random utility model. *Mathematical Social Sciences*, 43(3):303–343, 2002.